\documentclass[sn-mathphys]{sn-jnl}
\jyear{2021}%
\usepackage{graphicx}
\usepackage{subfloat}
\usepackage{subfig}
\usepackage{algpseudocode}
\usepackage{amssymb}
\usepackage{amsmath,amsfonts}
\usepackage{bm}
\usepackage{booktabs}
\usepackage{multirow}
\usepackage[switch]{lineno}  %
\usepackage[normalem]{ulem}
\useunder{ine}{}{}
\newcommand{\transp}{\mathsf{T}}
\begin{document}

\title[.]{Deep Federated Anomaly Detection for Multivariate Time Series Data}

\author[1]{\fnm{Wei} \sur{Zhu}}

\author[2]{\fnm{Dongjin} \sur{Song}}

\author[3]{\fnm{Yuncong} \sur{Chen}}
\author[3]{\fnm{Wei} \sur{Cheng}}
\author[3]{\fnm{Bo} \sur{Zong}}
\author[3]{\fnm{Takehiko} \sur{Mizoguchi}}
\author[3]{\fnm{Cristian} \sur{Lumezanu}}
\author[3]{\fnm{Haifeng} \sur{Chen}}
\author[1]{\fnm{Jiebo} \sur{Luo}}

\affil[1]{\orgname{University of Rochester}, \orgaddress{\state{NY}, \country{USA}}}

\affil[2]{\orgname{University of Connecticut}, \orgaddress{\state{CT}, \country{USA}}}

\affil[3]{\orgname{NEC American Lab}, \orgaddress{\state{NJ}, \country{USA}}}

\abstract{Despite the fact that many anomaly detection approaches have been developed for multivariate time series data, limited effort has been made on federated settings in which multivariate time series data are heterogeneously distributed among different edge devices while data sharing is prohibited. In this paper, we investigate the problem of federated unsupervised anomaly detection and present a Federated Exemplar-based Deep Neural Network (Fed-ExDNN) to conduct anomaly detection for multivariate time series data on different edge devices. Specifically, we first design an Exemplar-based Deep Neural network (ExDNN) to learn local time series representations based on their compatibility with an exemplar module which consists of hidden parameters learned to capture varieties of normal patterns on each edge device. Next, a constrained clustering mechanism (FedCC) is employed on the centralized server to align and aggregate the parameters of different local exemplar modules to obtain a unified global exemplar module. Finally, the global exemplar module is deployed together with a shared feature encoder to each edge device and anomaly detection is conducted by examining the compatibility of testing data to the exemplar module. Fed-ExDNN captures local normal time series patterns with ExDNN and aggregates these patterns by FedCC, and thus can handle the heterogeneous data distributed over different edge devices simultaneously. Thoroughly empirical studies on six public datasets show that ExDNN and Fed-ExDNN can outperform state-of-the-art anomaly detection algorithms and federated learning techniques.}


\maketitle

\section{Introduction}
\label{sec:introduction}

{A}{nomaly} detection in multivariate time series refers to identifying abnormal status in certain time steps of the time series data~\cite{kim2012robust,zong2018deep}. Building an effective unsupervised anomaly detection algorithm, however, is challenging since it requires collecting and profiling as much as (normal) multivariate time series data so as to reduce potential false positives~\cite{ruff2018deep,malhotra2015long,zhang2019deep}. With the rapid development of 5G networks, multivariate time series data are increasingly collected in various types of Internet of Things (IoT) edge devices, \textit{e.g.}, mobile phones, healthcare, wearable devices, \textit{etc}. However, due to privacy concerns~\cite{geyer2017differentially}, regulations~\cite{yang2019federated}, and transmission load~\cite{bonawitz2019towards}, directly transferring data from edge devices to a centralized server in order to train a unified anomaly detection model is usually infeasible or prohibited~\cite{yang2019federated,mcmahan2017communication}. Consequently, there is a huge demand to develop an anomaly detection algorithm that can collaboratively handle the multivariate time series data distributed on different edge devices while preserving privacy.

For this purpose, we resort to federated learning and aim to conduct privacy-preserving anomaly detection. Specifically, assuming the normal status of multivariate time series data consists of $K$ different modes which are heterogeneously distributed over $L$ different edge devices, we aim to learn a unified model that can not only preserve data privacy on each edge device but also identify all anomalous situations accurately. A notable challenge is that due to the environmental conditions or other external factors, the time series data collected on each edge device may only partially (\textit{i.e.}, less than $K$ modes) cover all different modes of normal status. In this case, simply training an anomaly detection model may involve many false positives. Taking wearable devices as an example, a mode can be ``walking", ``sitting", ``running", ``bicycling", or ``standing", while an anomaly can represent ``falling down" which is unusual to happen. A notable issue with this setting is that due to environmental conditions or other external factors, the time series data collected on each edge device may only partially cover the entire normal space. In other words, they are often heterogeneously distributed among different edge devices. For instance, an elder person tends to have activities of ``walking", ``sitting", and ``standing", while a young person may prefer activities of ``walking", ``running", and ``bicycling". In this case, if we only train an anomaly detection model based on the data collected from the elder person's edge device, we may falsely detect ``running" and ``bicycling" as anomalies (\textit{i.e.}, ``falling down") since they are not considered in the training.

Directly applying existing federated learning approaches to perform unsupervised anomaly detection in multivariate time series data often leads to inferior performance. This is because most existing anomaly detection methods assume that the entire normal space is covered by the training data~\cite{pang2020deep} which may not be true with the current setting; another reason is that existing federated learning algorithms, \textit{e.g.}, Federated Averaging (FedAvg)~\cite{mcmahan2017communication}, are originally designed for supervised learning and may not be able to properly handle the unsupervised tasks with heterogeneous data distributed on edge devices. Simply combining federated learning and unsupervised anomaly detection algorithms could lead to a series of problems. One issue is that each locally trained model may only partially cover the entire normal space, the federated learning algorithms, \textit{e.g.}, federated averaging operation, could produce a global model which may collapse to certain normal modes and even cover a certain part of the abnormal space~\cite{wang2020federated}. As a result, many anomalies could be mistreated as normal status and vice versa. Another issue is that existing unsupervised anomaly detection methods often rely on the encoder-decoder framework~\cite{malhotra2015long,zhang2019deep} and Generative Adversarial Network (GAN)~\cite{goodfellow2014generative,schlegl2017unsupervised} to extract semantic representation, and the extra parameters brought by the decoder and generator may result in a heavy communication cost.

To address the aforementioned issues, in this paper, we present a Federated Exemplar-based Deep Neural Network (Fed-ExDNN) to perform federated anomaly detection with multivariate time series data. On the edge device side, we specifically designed an Exemplar-based Deep Neural Network (ExDNN) to perform anomaly detection. ExDNN can simultaneously learn local time series representations based on their compatibility with an exemplar module 
which consists of hidden parameters learned
to capture varieties of normal patterns in the hidden feature space. On the server side, to cope with the heterogeneity of time series data on different edge devices, Fed-ExDNN employs a Federated Constrained Clustering (FedCC) technique to align and aggregate parameters of different local exemplar modules. Eventually, the updated global exemplar module together with a shared feature encoder will be sent back to edge devices and the anomaly detection is conducted by measuring the compatibility of test data (with extracted features) to the global exemplar module.


The main contributions of this paper are summarized as follows:
\begin{itemize}
    \item We formally investigate the problem of federated unsupervised anomaly detection (FedUAD) for multivariate time series data and develop Fed-ExDNN which consists of ExDNN for local anomaly detection and FedCC for model aggregation to handle FedUAD. 
    \item On the edge device side, Exemplar-based Deep Neural Network (ExDNN) can simultaneously learn local time series representations based on their compatibility with an exemplar module which is developed to capture potential normal patterns in the hidden feature space. On the server side, FedCC could align and aggregate different local exemplar modules. ExDNN and FedCC work jointly to address the heterogeneous distribution among edge devices.
    
    \item Our empirical studies on six public multivariate time series datasets demonstrate the effectiveness of the proposed ExDNN and Fed-ExDNN. 
\end{itemize}

\section{Related Work}
The proposed Fed-ExDNN is closely related to unsupervised anomaly detection and federated learning.
\subsection{Unsupervised Anomaly Detection}
Recently, deep learning-based anomaly detection methods has shown fruitful progress compared with traditional methods including one class SVM~\cite{scholkopf2000support}, Isolated Forest~\cite{liu2008isolation}, \textit{etc.}. In general, they can be categorized into four different types, \textit{i.e.}, One-class classification based methods, reconstruction based approaches, contrastive learning based techniques, and clustering based methods.

For one-class classification based methods, Deep Support Vector Data Descriptor (SVDD) replaces the kernel used in OCSVM~\cite{scholkopf2000support} with a deep neural network~\cite{ruff2018deep}. Context Vector Data Description (CVDD) generates multi semantic contexts by multi-head attention mechanism \cite{ruff2019self}. Temporal Hierarchical One-Class (THOC) conducts multi-scale one class learning in a hierarchical manner \cite{shen2020timeseries}. Reconstruction-based approaches mainly rely on autoencoder framework to reconstruct the input and employ the reconstruction error to detect anomalies. For instance, LSTM Autoencoder (AE)~\cite{malhotra2015long} adopts LSTM to encode and decode multivariate time series (MTS) data. To better model inter-correlation between different time series, Multiscale Convolutional Recurrent Encoder-Decoder (MSCRED) is proposed to reconstruct system signature matrices by an attention-based convLSTM~\cite{zhang2019deep}.  Memory Augmented Autoencoder (MemAE) augments autoencoder with an external memory~\cite{gong2019memorizing}. BeatGAN~\cite{zhou2019beatgan} regularizes the reconstructed data by a generative adversarial network~\cite{goodfellow2014generative}. OmniAnomaly learns robust representation with stochastic variable connection and planar normalizing flow~\cite{su2019robust}. Unsupervised Anomaly Detection (USAD) imposes additional constraints on reconstruction by an additional decoder \cite{audibert2020usad}. Li \textit{et al.} conduct anomaly detection with hierarchical VAE and low dimensional embedding for anomaly detection~\cite{li2021multivariate}.  More recently, contrastive learning based techniques are becoming popular for time series anomaly detection. For instance, self-supervised contrastive predictive coding is proposed to handle anomaly points~\cite{deldari2021time}. Cho \textit{et al.} propose a masked contrastive method by using class-wise scale factor~\cite{cho2021masked}. A unified contrastive anomaly detection framework is proposed by~\cite{sehwag2021ssd}.  Carmona \textit{et al.} perform time series anomaly detection by generating abnormal series with expertise knowledge~\cite{carmona2021neural}. Qiu \textit{et al.} propose deterministic contrastive loss to enable the anomaly score to be consistent with training loss~\cite{qiu2021neural}. Clustering-based deep neural networks are also applied to anomaly detection. For instance, Zong \textit{et al.} proposes Data Encoder Gaussian Mixture Model (DAGMM) for anomaly detection and they conduct GMM on the feature space composed of reconstruction score and encoding of the autoencoder.  In this paper, we introduce a novel task as Federated Unsupervised Anomaly Detection (FedUAD). The heterogeneous data distribution on edge devices and data-free server-side model aggregation brings challenges for existing anomaly detection methods. We specially design Fed-ExDNN to overcome these problems.


\subsection{Federated Learning}
Federated learning receives increasing attention recently~\cite{yang2019federated,geyer2017differentially,konevcny2016federated,liu2021fedlearn}. One of the major concerns of federated learning is the heterogeneity problem~\cite{sahu2018convergence,li2019convergence,karimireddy2020scaffold,reisizadeh2020robust,dinh2020personalized,zhu2021data}. Federated Averaging (FedAvg) is the most widely used algorithm for federated tasks~\cite{li2019convergence}. Federated Proximal (FedProx) is proposed to alleviate the heterogeneity challenge~\cite{sahu2018convergence}. Sattler \textit{et al.} (\cite{sattler2019clustered}) propose to hierarchically cluster the locally learned models. Xie \textit{et al.} (\cite{xie2020multi}) propose to maintain multi global models and assign user's gradient to different global models.  Liang \textit{et al.}~(\cite{liang2020think}) propose to learn local representations on each device and an overall global model across devices . Federated Matching Average (FedMA) learns to cluster the local models before averaging the weights \cite{wang2020federated}. Yu \textit{et al.}~(\cite{yu2020federated}) propose FedAwS where there is only one class on each device. Fedfast is proposed to handle federated recommendation system \cite{muhammad2020fedfast}. FedDF conducts data-free knowledge distillation on server side to aggregate local models\cite{lin2020ensemble}. Fallah \textit{et. al.}~(\cite{fallah2020personalized}) applies meta learning \cite{finn2017model} on client updates. However, these methods are all proposed to handle supervised federated learning tasks. The unsupervised setting of UAD makes the heterogeneity problem much more challenging, and we find that the proposed Fed-ExDNN has clear advantages over conventional federated learning methods for FedUAD. 

More recently, federated anomaly detection draws increasing attention~\cite{singh2021anomaly,nguyen2019diot,zhao2019multi,sater2021federated,wang2021towards}. Nguyen \textit{et al.} propose to apply federated anomaly detection for IoT devices~\cite{nguyen2019diot}. Federated anomaly detection is also used to address the IoT security attacks~\cite{mothukuri2021federated}. Liu \textit{et al.} propose an on-device method for industrial IoT anomaly detection. Zhao \textit{et al.} propose a multi-task network for federated anomaly detection~\cite{zhao2019multi}. Compared to existing works, our proposed Fed-ExDNN focuses on unsupervised anomaly detection and employs an effective method to address the heterogeneity problem.



\section{Federated Exemplar-based Deep Neural Network}
 In this section, we present a Federated Exemplar-based Deep Neural Network (Fed-ExDNN) to perform federated unsupervised anomaly detection. Fed-ExDNN consists  of  ExDNN  for  local anomaly  detection  and Federated Constrained Clustering  (FedCC) for model aggregation.
 
We find that simply combining existing federated learning and anomaly detection approaches to handle the task often leads to inferior performance. This should be attributed to the fact that most existing anomaly detection methods are developed based on the assumption that the entire normal space is covered by the training data~\cite{pang2020deep}, and this assumption does not hold when we conduct training on edge devices with heterogeneously distributed data, \textit{i.e.}, not all normal patterns can be accessed during the training stage on each edge device. On the other hand, existing federated learning algorithms, \textit{e.g.}, Federated Averaging (FedAvg)~\cite{mcmahan2017communication}, are originally designed for supervised classification tasks and they may not properly aggregate the unsupervised anomaly detection models trained on edge devices. To this end, we present Fed-ExDNN, which consists of a clustering-based anomaly detection method ExDNN and a federated aggregation method FedCC to align and aggregate different local exemplar modules.

\begin{figure*}
\centering
\includegraphics[width=0.9\textwidth]{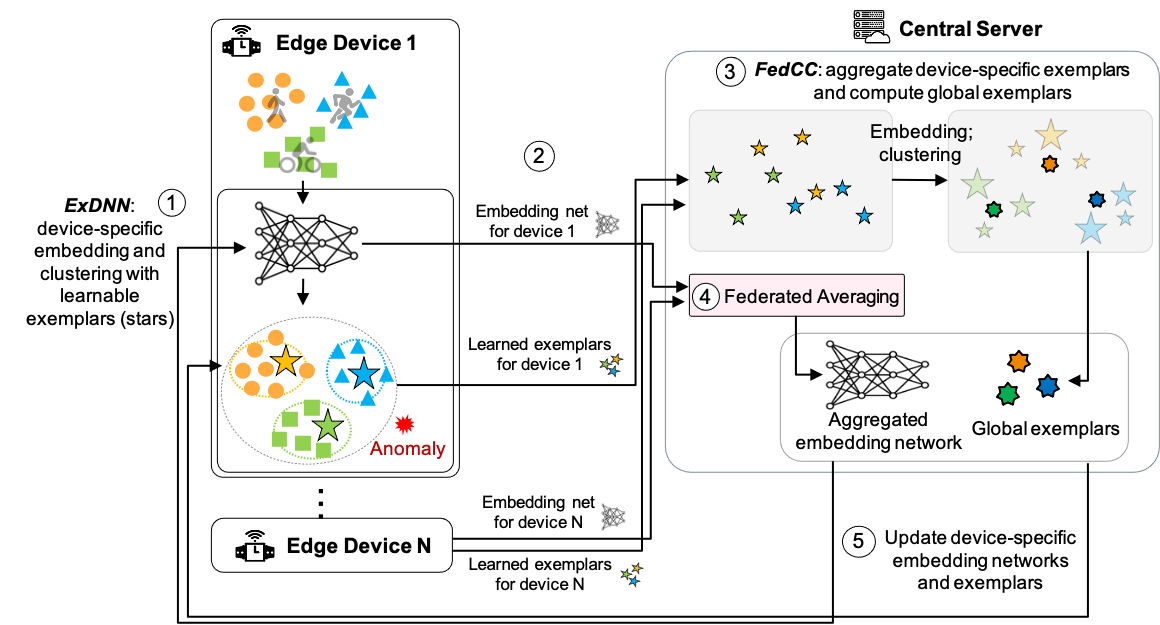}
\caption{Block diagram of the proposed Fed-ExDNN which contains five steps for each communication round. (1) We train the device-specific model, including the embedding network and learnable local exemplars (shown as stars), on each edge device by the proposed ExDNN. The models on all devices are initialized with a global model fetched from the central server. (2) Edge devices send device-specific embedding networks and learned local exemplars to the server. (3) Compute global exemplars from the local exemplars using the proposed Federated Constrained Clustering (FCC). (4) Aggregate the local models to obtain the global model using FedAvg. (5) Updated global model is sent back to edge devices. The procedure will be repeated until convergence is achieved.}
\label{fig:diag}
\end{figure*}
 
 We brief the basic training procedure as follows: 
 assuming there are $L$ edge devices, the $l$-th local device learns a device-specific model, which includes an embedding network $f^l(\cdot; \theta^l)$ for feature encoding and an exemplar module in which a set of $K$ local exemplars $\mathbf{C}^l = \{\mathbf{c}_1^l, \cdots, \mathbf{c}_K^l  \}\in\mathbb{R}^{d\times K}$ is learned to capture the potential normal patterns in the hidden feature space. The local model is trained based on the time series data collected on the $l$-th local device for unsupervised anomaly detection. The central server aggregates local models from different devices to construct a global model. The embedding network (for feature encoding) of the global model $g(\cdot; \bar{\theta})$ is obtained by Federated Averaging, and the global exemplar module with $K$ learnable exemplars $\mathbf{U} = \{\mathbf{u}_1, \cdots, \mathbf{u}_K  \}$ is obtained by aggregating and align all local exemplar modules with the proposed FedCC. Finally, the server sends the global model to different edge devices to update their local models. Fed-ExDNN jointly adopts FedCC and ExDNN to enable the global model to capture entire normal patterns even when the data are heterogeneously distributed among edge devices. Please refer to Fig. \ref{fig:diag} for detailed illustration.

\subsection{Local Device: ExDNN}
\begin{figure}
\centering
\includegraphics[width=0.8\textwidth]{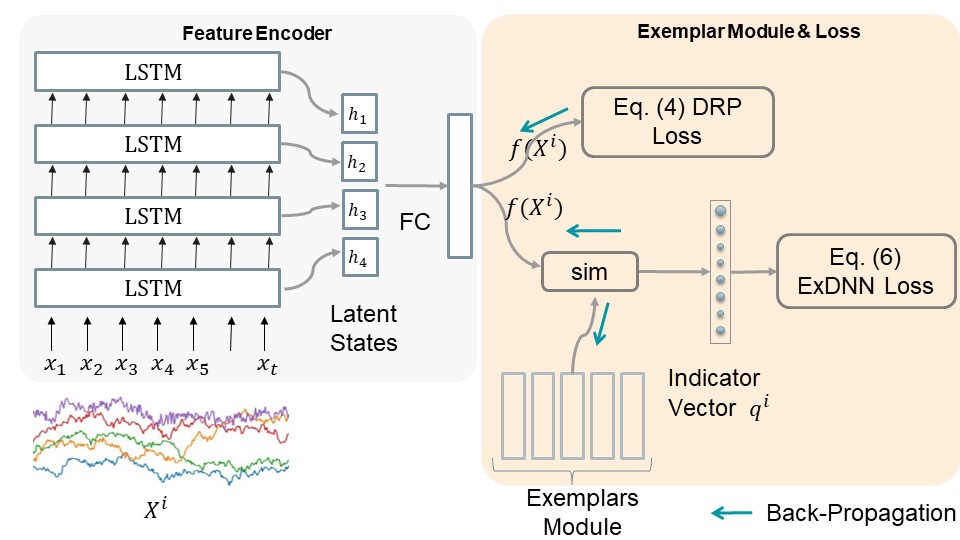}
\caption{The time series segment $X^i$ is processed by a 4-layer LSTM followed by a fully connected embedding layer. The exemplars module are updated by gradient descent and is implemented with a fully connected layer.}
\label{fig:localexdnn}
\end{figure}

Exemplar-based Deep Neural Network (ExDNN), as a clustering-based anomaly detection method, is naturally suitable for handling heterogeneous data. 
Compared with existing clustering-based methods \cite{liao2018unified,zong2018deep} that adopt Gaussian Mixture Model (GMM) as the clustering objective, the ExDNN is developed based on an advanced clustering algorithm \cite{hu2017learning,xie2016unsupervised}. Moreover, we propose Deep Relation Preserving (DRP) to learn the representation of multivariate time series data in an unsupervised manner. We'd like to emphasize that ExDNN conducts clustering and representation learning simultaneously to mutually boost their performance for better anomaly detection.  The details of ExDNN are shown in Fig. \ref{fig:localexdnn}.  In the following, the algorithm described is for one particular local device, so we omit the superscript $l$ for brevity.

For a specific local device with $n$ multivariate time series segments $\{\mathbf{X}^{i}\}_{i=1}^{n}\in\mathbb{R}^{m\times t}$, where $m$ denotes the number of time series and $t$ is the length of a segment, our method learns the optimal embedding network $f(\cdot; \theta)$ and an exemplar module with a learnable parameter $\mathbf{C}\in\mathbb{R}^{d\times K}$. In this paper, we use LSTMs to encode temporal dynamics in the multivariate time series. The exemplar module $\mathbf{C}$ is implemented by a fully connected layer, and is jointly trained with the embedding network parameter $\theta$ in a end-to-end manner.

Specifically, our proposed method is motivated by Deep Embedding Clustering \cite{xie2016unsupervised} with the following objective: 
\begin{equation} \label{eq:localcluster}
\begin{split}
    \min_{\theta, \mathbf{C}} \frac{1}{n} \sum_{i=1}^n KL (\mathbf{p}_i\Vert \mathbf{q}_i)
\end{split}
\end{equation}
where $KL(\cdot \Vert \cdot)$ is the Kullback–Leibler divergence and Eq. (\ref{eq:localcluster}) encourages the exemplars to be close to the training samples on the embedding space, and each learned exemplar could then capture a specific pattern of the normal data just like a clustering center in K-means. Specifically, $\mathbf{q}_i\in R^{K}$ is the cluster indicator vector for the $i$-th segment where $q_{ij}$ is the probability of assigning the $i$-th data to the $j$-th exemplar. This probability is computed as
\begin{equation} \label{eq:q}
    \begin{split}
        q_{ij} = \frac{\exp(\gamma_1 s(f(\mathbf{X}^{i}), \mathbf{c}_j))}{\sum_{k=1}^K \exp(\gamma_1 s(f(\mathbf{X}^{i}), \mathbf{c}_k))}, 
    \end{split}
\end{equation}
where $s$ is the cosine similarity $s(\mathbf{a}, \mathbf{b}) = \frac{\mathbf{a}^\transp \mathbf{b}}{\Vert \mathbf{a} \Vert_2 \Vert \mathbf{b}\Vert_2}$, $\gamma_1$ is a learnable scale factor, and $\mathbf{C} = \{\mathbf{c}_1, \cdots, \mathbf{c}_K  \}\in\mathbb{R}^{d\times K}$ denote local exemplars. We should highlight that the local exemplars are the same as the weights of a fully connected layer and their privacy concern could be addressed by well-studied approaches, \textit{e.g.}, differential privacy \cite{geyer2017differentially}. Following \cite{xie2016unsupervised}, we raise $\mathbf{q}_i$ to the second power and normalize it by the size of clusters to obtain $\mathbf{p}_i$ as
\begin{equation} \label{eq:p}
    p_{ij} = \frac{q_{ij}^2 / \sum_{i'=1}^n q_{i'j}}{\sum_{j'} (q_{ij'}^2/\sum_{i'=1}^n q_{i'j'})}
\end{equation}

However, exemplar learned by optimizing Eq. (\ref{eq:localcluster}) may converge to patterns with few samples and even noisy data, and we then adopt the balanced loss to alleviate the problem as
\begin{equation} \label{eq:localbalance}
\begin{split}
    \min_{\theta, \mathbf{C}}  - \bm{\alpha}^\transp \log\left(\frac{1}{n}\sum_{i=1}^n \mathbf{q}_i \right)
\end{split}
\end{equation}
$\bm{\alpha} \in \mathbb{R}^{K}$ is a prior distribution over the exemplars, and this term encourages the cluster sizes on edge devices to match the prior \cite{hu2017learning}.
We set $\bm{\alpha} = \frac{1}{K} \textbf{1}$, \textit{i.e.}, uniform distribution. The effectiveness of the balancing term on handling contaminated data is shown in our ablation studies.


Representation learning is critical for deep neural network training, and DEC is often trained with auto-encoder initialization \cite{xie2016unsupervised}, which may lead to privacy concern and computational cost especially in federated learning settings.
Furthermore, we propose Deep Relative Preserving (DRP) that encourages the latent space to preserve the local similarity induced by the original feature\cite{roweis2000nonlinear,chui2018deep}:
\begin{equation} \label{eq:localmetric}
\begin{split}
M(\mathbf{X}^i) = 
\min_\theta \log\left(1  + \sum_{\substack{p\in \mathcal{P}_i \\ n \notin \mathcal{P}_i}} \exp(\gamma_2 (s_{in}-s_{ip})) \right) 
\end{split}
\end{equation}
where $\mathcal{P}_i$ is the set of nearest neighbors of the $i$-th example.
$\gamma_2$ is a learnable scale factor. $s_{ij}$ is shorthand for
$s(f(\mathbf{X}^{i}), f(\mathbf{X}^{j}))$. Eq. (\ref{eq:localmetric}) encourages the similarity of positive pairs to be larger than that of negative pairs \cite{sun2020circle}. Moreover, to avoid computation and storage cost of the KNN graph, we approximate KNN by the samples within each minibatch.

We could perform anomaly detection by jointly optimizing Eq. (\ref{eq:localcluster}), Eq. (\ref{eq:localbalance}), and Eq. (\ref{eq:localmetric}) as
\begin{equation} \label{eq:localall1}
\begin{split}
    \min_{\theta, \mathbf{C}} \frac{1}{n} \sum_{i=1}^n \left( KL (\mathbf{p}_i\Vert \mathbf{q}_i)+M(\mathbf{X}^i)\right)- \bm{\alpha}^\transp \log\left(\frac{1}{n}\sum_{i=1}^n \mathbf{q}_i \right),
\end{split}
\end{equation}
and the \textit{anomaly score} for sample $\mathbf{X}$ is calculated by the negative cosine similarity between the samples and its nearest exemplars as
\begin{equation}
\label{eq:anomalyscore}
\begin{split}
\text{Score}(\mathbf{x}) = - \max_{j} \; s(f(\mathbf{X}), \mathbf{c}_j).
\end{split}
\end{equation}
However, Eq. (\ref{eq:localall1}) only forces the relative similarity between the sample and its nearest exemplar to be larger than the similarity between the sample and other exemplars \cite{xie2016unsupervised}. After training, we actually have little guarantee on the numeric value of anomaly score. Therefore, although Eq. (\ref{eq:localall1}) may be effective for clustering, it leads to sub-optimal performance for anomaly detection in our experiments. To alleviate the problem, we introduce an absolute term to directly optimize the numeric value of anomaly score, and our final objective for anomaly detection on a local device becomes:
\begin{equation} \label{eq:localall}
\begin{split}
\min_{\theta, \mathbf{C}}  \frac{1}{n} \sum_{i=1}^n  \left(KL (\mathbf{p}_i\Vert \mathbf{q}_i)+M(\mathbf{X}^i)\right) - \bm{\alpha}^\transp \log\left(\frac{1}{n}\sum_{i=1}^n \mathbf{q}_i \right) \\
+ \frac{1}{n} \sum_{i=1}^n \log\left(1+\exp(-\gamma_3 ( s(f(\mathbf{X}^i), \Bar{\mathbf{c}}_i) -m) \right).
\end{split}
\end{equation}
The last term of Eq. (\ref{eq:localall}) \footnote{Here we slightly abuse the symbol of cosine similarity and $s(f(\mathbf{X}^i), \Bar{\mathbf{c}}_i)$ is calculated by the dot product between $f(\mathbf{X}^i)^T$ and $\Bar{\mathbf{c}}$ with normalized $f(\mathbf{X}^i)$ and $\mathbf{c}_j$.} is the proposed absolute score loss. Since the max operation in Eq. (\ref{eq:anomalyscore}) is non-differentiable, our absolute score term maximizes the cosine similarity between the $i$-th segment and a soft approximation of the nearest exemplar $\Bar{\mathbf{c}}_i = \sum_{j=1}^K {q}_{ij}\mathbf{c}_j$, where $\mathbf{q}_i$ is the soft indicator vector for the $i$-th sample. $\gamma_3$ is a learnable scale factor and $m>0$ is the margin.We adopt softplus operation to make the proposed term have similar scale as other terms. Since the distribution of pairwise cosine similarity of two random high-dimensional unit vectors approaches a zero mean Gaussian, it is thus necessary to have $s(f(\mathbf{X}^i), \Bar{\mathbf{c}}_i)$ larger than a positive margin $m$. We highlight that although the ExDNN in Eq. (\ref{eq:localall}) contains several hyper-parameters, most of them are fixed during the empirical studies in this paper besides the number of exemplars $K$.


After certain rounds of training on the $l$-th local device, the parameters of the embedding network $\theta^l$ and the set of exemplars $\{\mathbf{c}_1^l, \cdots, \mathbf{c}_K^l  \}$ are all uploaded to a central server.

\subsection{Central Server: FedCC for Exemplars Aggregation}
The central server aggregates the local models uploaded from edge devices to obtain a global feature encoding network $g$ and a global exemplar module $\mathbf{U}=\{\mathbf{u}_1, \cdots, \mathbf{u}_K  \}$ that capture the heterogeneous data distribution on all edge devices. The feature encoding network could be aggregated by existing federated learning methods, \textit{e.g.},  federated averaging (FedAvg) \cite{mcmahan2017communication} and Federated Proximal (FedProx) \cite{sahu2018convergence}. However, due to heterogeneous data on edge devices, the local exemplar module, even with the same initialization, may significantly deviate from each other to better fit the local data. Even worse, due to the cost of transmission, the local training is desired to be longer to reduce the communication load between server and clients. As a result, the alignment between the updated local exemplar module and the previous global exemplar module may not hold as the training on local devices proceeds.

Most of existing federated learning methods, \textit{e.g.}, FedAvg and FedProx, element-wisely averages the local exemplars based on the assumption that all local exemplar modules and the global exemplar module are still well-aligned during the training. For supervised tasks, the alignment could be regularized by the ground truth label. For the proposed Federated Unsupervised Anomaly Detection (FedUAD) task, however, no label is available. Therefore, vanilla federated learning methods will lead to suboptimial performance in practice.
Aggregating the exemplar modules by K-means seems to be a reasonable choice \cite{wang2020federated}. However, K-means may also results in misalignment since its objective and the representation is decoupled which make it impossible to adjust the representation so as to mitigate the issue. In this paper, we propose an innovative approach, namely Federated Constrained Clustering (FedCC) to address the heterogeneity problem. The basic idea is to first learn a projection function $h$ that could align the local exemplars to discover and enhance the alignment, and then conduct clustering on the learned embedding space to obtain the global exemplar module. The effectiveness of FedCC is verified in the experiment section.


From $L$ edge devices each with exemplar module consisting of $K$ learnable exemplars, the central server receives a total of $N=LK$ local exemplars  which is denoted as $\{\mathbf{c}_1^1, \cdots, \mathbf{c}_K^1, \mathbf{c}_1^2, \cdots, $ $\mathbf{c}_K^2, \cdots, \mathbf{c}_1^L \cdots \mathbf{c}_K^L  \}$.
The proposed FedCC is formulated as:
\begin{equation} \label{eq:server}
\begin{split}
\min_{\phi, \{\mathbf{v}_1, \cdots, \mathbf{v}_K\}} -\frac{1}{N}\sum_{i=1}^{K}\sum_{l=1}^L \mathbf{p}_{il}^\transp \log \mathbf{q}_{il} \\
- \textbf{1}^\transp  \log \left(\frac{1}{N}\sum_{i=1}^{K} \sum_{l=1}^L \mathbf{p}_{il} \right) + \frac{1}{N}\sum_{i=1}^K\sum_{l=1}^L R(\bm{c}_i^l),
\end{split}
\end{equation}
where $\phi$ denotes the parameters of the projection network $h$, and $\{\mathbf{v}_1, \cdots, \mathbf{v}_K\}$ are the latent cluster centers in the output space of $h$.
The first two terms in Eq. (\ref{eq:server}) are for clustering, similar to the first and third terms in Eq. (\ref{eq:localall}).
$\mathbf{q}_{il}$ are defined similarly to Eq. (\ref{eq:q}):
\begin{equation} \label{eq:q_global}
    \begin{split}
        q_{il,j} = \frac{\exp(\gamma_4 \cdot s(h(\mathbf{c}_i^l), \mathbf{v}_j))}{\sum_{k=1}^K \exp(\gamma_4 \cdot s(h(\mathbf{c}_i^l), \mathbf{v}_k))}, 
    \end{split}
\end{equation}
and $\mathbf{p}_{il}$ is defined based on $\mathbf{q}_{il}$ in the same way as Eq. (\ref{eq:p}), and $\gamma_4$ is the scale factor.
 We further introduce the constraints $R$ to encourage the learned projection $h$ to give similar embedding for exemplars that have the same initialization (stars with same color in Fig. \ref{fig:diag}).
\begin{equation}
\begin{split}
    R(\mathbf{c}_i^l) = \log \left( 1+ 
    \sum_{m=1}^L \sum_{j=1}^K \exp(\gamma_5 \mathbf{e}_{ij}  s(h(\mathbf{c}_i^l),h(\mathbf{c}_j^m))) \right),
\end{split}
\end{equation}
where $\mathbf{e}_{ij} = 1$ if $i = j$ and $c_i^l$ is K nearest neighbor of $c_j^m$, and $\mathbf{e}_{ij} = -1$ otherwise. $\gamma_5$ is a learnable scale factor.
Finally, the global exemplar module $\{\mathbf{u}_1, \cdots, \mathbf{u}_K\}$ can be obtained based on the clustering indicator matrix:
\begin{equation}
    \mathbf{u}_z = \frac{1}{\sum_{i=1}^{K} \sum_{l=1}^L q_{il,z}}\sum_{i=1}^{K} \sum_{l=1}^L q_{il,z} \mathbf{c}_i^l.
\end{equation}
After obtaining the global exemplar module and the averaged embedding network (for feature encoding), they will be send back to each edge devices for the next round of learning.  

\section{Experiments}

In this section, we verify the superiority of ExDNN and Fed-ExDNN for anomaly detection. 
\subsection{Datasets and Evaluation Metrics}
We conduct experiments on six publicly available multivariate time series datasets, including 2D Gesture~\cite{keogh2005hot}, ECG5000~\cite{goldberger2000physiobank}, SWaT~\cite{mathur2016swat}, HAR Laying~\cite{anguita2013public}, UWave~\cite{liu2009uwave}, and ArabicDigits~\cite{hammami2009tree}. The details of these datasets and train/validation/test partitions are summarized as follows and in Table \ref{tab:data}.
\begin{itemize}
    \item 2D Gesture records the X-Y coordinates of a hand gesture in a video \cite{keogh2005hot};
    \item SWaT contains the data of a water treatment plant for water filtering. The data are composed of 11 days of operations with 7 days normal data and 4 days attack data \cite{mathur2016swat}; 
    \item ECG5000 records ECG series of a patient who has sever congestive heart failure \cite{goldberger2000physiobank};
    \item HAR Laying is a human activity recognition dataset that contains 6 different activities performed by 30 subjects including walking, walking upstairs, walking downstairs, sitting, standing, and laying. The dataset has 30 subjects with 21 as training and 9 as testing. The laying samples are regarded as anomaly cases. We select 5 subjects from training data for validation \cite{anguita2013public}. 
    \item AerobicDigit \cite{hammami2010improved} contains Mel-Frequency Cepstrum Coefficient (MFCC) data of spoken arabic digits, and each digit is alternatively regarded as anomaly following the setting in \cite{ruff2018deep}. 
    \item UWave \cite{liu2009uwave} consists of x, y, z coordinates of eight gestures recorded by accelerometers of Wii remotes, and each gesture is alternatively regarded as anomaly following the setting in  \cite{ruff2018deep}. 
\end{itemize}

\begin{table}[]
\centering
\caption{The detailed statistics of six multivariate time series datasets.}
\label{tab:data}
\resizebox{0.6\columnwidth}{!}{
\begin{tabular}{@{}llcccc@{}}
\toprule
Dataset       & \# train - val - test & \# dim ($m$) & \# Length ($t$) \\ \midrule
2D Gesture    & 8171 - 876 - 2044       & 2                  & 80        \\
SWaT          & 47420 - 11198 - 33594    & 51                 & 100       \\
ECG5000       & 292 - 1125 - 3375       & 1                  & 140       \\
HAR Laying    & 4559 - 1676 - 2947       & 9                  & 128       \\
AerobicDigits & 6600 - n/a - 2200       & 13                 & 20        \\ 
UWave         & 1600 - n/a - 2879       & 3                  & 40        \\ \bottomrule
\end{tabular}
}
\end{table}

For 2D Gesture and SWaT, we use sliding window with stride 1 to partition them. Moreover, since their training sets only have normal data, we select a portion of data from original testing data to construct the validation set for hyper-parameter tuning. We downsample the time series of SWaT by 10 following \cite{shen2020timeseries}. The length of segments of AerobicDigit and Uwave varies and we resize their segments to 20 and 40 respectively. For 2D Gesture, ECG5000, SWaT, and HAR Laying, we set hyperparameters based on grid search over the validation set. We report the AUC, F1, Precision and Recall linked to the best F1 score on validation set following~\cite{audibert2020usad}. For AerobicDigit and UWave, following the setting similar to \cite{ruff2018deep}, we do not construct validation set, and only report average AUC by iteratively treating each class as the anomaly case. All experiments are ran three times and mean and standard deviations of all metrics are provided.


\subsection{ExDNN for Anomaly Detection} \label{sec:expexdnn}

We first conduct experiments to show the effectiveness of the proposed ExDNN for unsupervised anomaly detection with multivariate time series (MTS) data.

\subsubsection{Comparison Methods and Experimental Settings} \label{sec:expsetting}
We compare the proposed ExDNN with seven deep learning methods, including LSTM-AutoEncoder (LSTM-AE) \cite{malhotra2015long}, BeatGAN \cite{zhou2019beatgan}, Memory Augmented AutoEncoder (MemAE) \cite{malhotra2015long}, USAD \cite{audibert2020usad}, Deep SVDD \cite{ruff2018deep},  Contextual SVDD (CVDD) \cite{ruff2019self}, and Deep Autoencoding Gaussian Mixture Model (DAGMM) \cite{zong2018deep}. Among them, LSTM AE, BeatGAN, MemAE, and USAD are reconstruction-based methods, while Deep SVDD and CVDD are one class-based methods. DAGMM jointly uses auto-encoder reconstruction and clustering for anomaly detection.

All deep learning methods are implemented with the same feature encoder shown in Fig. \ref{fig:localexdnn}, which is composed of a 4-layer LSTM with the hidden dimension set as $8$. The decoder for reconstruction-based methods is with the same structure as the encoder. We fix the batch size as 128 and fix the learning rate set as 0.005 for all methods. We conduct an exhaustive grid search for the deep learning methods to find the optimal parameters for each dataset. For ExDNN, we fix the hyperparameter as  $\gamma_1 = 2$, $\gamma_2 = 10$, $\gamma_3 = 10$, $m=0.5$, and the number of nearest neighbors in DRP as 10, unless otherwise stated. We search the number of clusters from $\lbrace 8,16,32,64,128 \rbrace$. Since AerobicDigit and UWave do not have a validation set, we set the number of exemplars as 32 and 64 respectively. All experiments are conducted on a server with 4 Nvidia GTX 2080 Ti graphics cards. We summarize all methods as follows:
The detail settings of compared methods are shown as follows:
\begin{itemize}
    \item LSTM-AutoEncoder (LSTM-AE) is implemented with 4-layer BiLSTM as encoder and decoder \cite{malhotra2015long}.
    \item Deep SVDD is implemented also with a 4-layer BiLSTM as feature extractor. We first pretrain the network with LSTM-AE, and then train Deep SVDD by searching the $\lambda$ from $\lbrace 10^{-5},10^{-4}, 10^{-3}, 10^{-2}, 10^{-1} \rbrace$ \cite{ruff2018deep}.
    \item Memory Augmented AutoEncoder (MemAE) adopts similar network structure as LSTM-AE. We search the memory size from $\lbrace 32, 64, 128, 256 \rbrace$ and $\alpha$ from $\lbrace 10^{-3},10^{-2},10^{-1},1,10,100 \rbrace$ \cite{gong2019memorizing}. 
    \item Contextual SVDD (CVDD) is an extension of SVDD by adopting multi context projection. We search the number of contexts from $\lbrace 3,5,10,32 \rbrace$ and scale factor $\alpha$ from $\lbrace 10^{-5},10^{-4},10^{-3},10^{-2},10^{-1} \rbrace$ \cite{ruff2019self}.
    \item Deep Autoencoding Gaussian Mixture Model (DAGMM) is also built on the LSTM-AE and adopts a gaussian mixture model on the features composed of reconstruction loss and latent embedding of autoencoder. We search $\lambda_1$ from $\lbrace 10^{-3}, 10^{-2},10^{-1},1,100,1000 \rbrace$ and   $\lambda_2$ from $\lbrace 10^{-5},10^{-4},10^{-3},10^{-2},10^{-1},1 \rbrace$ \cite{zong2018deep}.
    \item USAD \cite{audibert2020usad} is based on autoencoder framework. We search the hyper-paramter $\alpha$ from $\lbrace 0,0.1,0.2,0.5,0.7,0.9 \rbrace$.
    \item BeatGAN \cite{zhou2019beatgan} applies GAN for auto-encoder. We varies the $\Lambda$ from $\lbrace 10^{-3},10^{-2},10^{-1},1,100,1000 \rbrace$.
    \item ExDNN is proposed in this paper. We fix the hyperparameter as  $\gamma_1 = 2$, $\gamma_2 = 10$, $\gamma_3 = 10$, $m=0.5$, and the number of nearest neighbors in DRP as 10, unless otherwise stated. We search the number of clusters from $\lbrace 8,16,32,64,128 \rbrace$.
\end{itemize}

\subsubsection{Effectiveness of ExDNN}
The anomaly detection results on MTS datasets are shown in Table \ref{tab:exdnn4}. According to the results, several interesting points are summarized as follows. First, either reconstruction and one-class based methods cannot handle all different scenarios. By imposing additional restrictions on the latent space learned by auto-encoder, BeatGAN, MemAE, and USAD could potentially boost the performance of LSTM AE on homogeneous datasets especially on HAR Laying, but are of little use and even also degrade the performance on more heterogeneous datasets, \textit{e.g.}, SWaT and AerobicDigits. By contrast, Deep SVDD works much better for homogeneous datasets, \textit{e.g.} HAR Laying, but also suffers from the under-fitting problem with a lower recall score. CVDD could largely alleviate the under-fitting problem of SVDD by the multi-context learning. Second, the proposed ExDNN, although is designed for anomaly detection with heterogeneous normal samples, could handle the over-fitting and under-fitting problem by tuning the number of exemplars and generally achieves superior performance on all datasets. ExDNN outperforms compared methods, especially on datasets with heterogeneous normal cases. For example, ExDNN achieves 4.58\% improvements (regarding averaged AUC) on AerobicDigit compared with the second best method CVDD. This is because ExDNN not only extracts superior representations by DRP but also conducts effective deep clustering to generate representative exemplars for anomaly detection.

\begin{table*}[]
\centering
\caption{Anomaly detection performance on six MTS datasets. Best methods are highlighted in bold.}
\label{tab:exdnn4}
\resizebox{\textwidth}{!}{
\begin{tabular}{llcccccccc}
\hline
Dataset                     & Metric & LSTM AE     & BeatGAN    & MemAE      & USAD       & Deep SVDD  & CVDD       & DAGMM      & ExDNN(Ours) \\ \hline
\multirow{4}{*}{2D Gesture} & AUC    & 80.19$\pm$4.32 & 81.87$\pm$3.84 & 80.88$\pm$0.40 & 80.15$\pm$3.99 & 73.58$\pm$2.73 & \underline{82.64}$\pm$0.05 & 74.74$\pm$8.51 & \textbf{88.37}$\pm$0.82  \\
                            & F1     & 58.57$\pm$3.59  & 60.81$\pm$4.05 & \underline{62.77}$\pm$1.14 & 61.58$\pm$3.36 & 54.42$\pm$3.83 & 62.54$\pm$2.47 & 56.06$\pm$9.12 & \textbf{71.13}$\pm$2.31  \\
                            & Prec   & 50.28$\pm$2.05  & 58.10$\pm$8.94 & 57.84$\pm$3.85 & 55.84$\pm$0.73 & 49.98$\pm$8.83 & 62.88$\pm$5.16 & 47.54$\pm$8.69 & 66.96$\pm$5.74  \\
                            & Rec    & 70.66$\pm$8.15  & 65.83$\pm$7.05 & 69.05$\pm$2.86 & 68.79$\pm$7.37 & 63.38$\pm$9.15 & 62.48$\pm$1.25 & 71.82$\pm$1.47 & 76.40$\pm$2.39  \\ \midrule
\multirow{4}{*}{SWaT}       & AUC    & \underline{91.87}$\pm$0.85  & 91.39$\pm$0.30 & 91.29$\pm$0.47 & 90.69$\pm$0.33 & 89.51$\pm$8.02 & \textbf{94.70}$\pm$0.94 & 90.02$\pm$0.99 & 91.61$\pm$0.78  \\
                            & F1     & 81.89$\pm$1.99  & 80.65$\pm$1.70 & 81.52$\pm$0.52 & 80.25$\pm$0.48 & 79.86$\pm$5.55 & \underline{83.34}$\pm$1.63 & 80.07$\pm$3.50 & \textbf{87.34}$\pm$0.61  \\
                            & Prec   & 72.56$\pm$4.63  & 70.15$\pm$4.25 & 71.35$\pm$1.42 & 69.35$\pm$1.28 & 69.19$\pm$4.78 & 71.71$\pm$2.39 & 70.15$\pm$5.71 & 85.10$\pm$0.72  \\
                            & Rec    & 94.44$\pm$3.00  & 95.31$\pm$3.16 & 95.19$\pm$2.40 & 95.26$\pm$1.05 & 94.46$\pm$6.95 & 99.51$\pm$0.15 & 93.28$\pm$1.64 & 89.71$\pm$0.60  \\ \midrule
\multirow{4}{*}{ECG5000}    & AUC    & 95.04$\pm$0.63  & 94.85$\pm$0.61 & 94.88$\pm$1.24 & 93.73$\pm$0.82 & 90.35$\pm$5.40 & \underline{97.38}$\pm$0.25 & 89.14$\pm$0.99 & \textbf{98.45}$\pm$1.05  \\
                            & F1     & 92.09$\pm$1.57  & \underline{92.31}$\pm$0.16 & 92.16$\pm$0.99 & 91.69$\pm$1.48 & 85.91$\pm$7.74 & 91.20$\pm$0.26 & 88.00$\pm$1.43 & \textbf{92.86}$\pm$2.64  \\
                            & Prec   & 88.36$\pm$3.32  & 87.95$\pm$0.51 & 88.05$\pm$2.28 & 86.53$\pm$2.66 & 89.58$\pm$6.12 & 90.69$\pm$0.92 & 79.55$\pm$2.20 & 91.29$\pm$3.64  \\
                            & Rec    & 96.24$\pm$0.85  & 97.12$\pm$0.88 & 96.74$\pm$0.85 & 97.57$\pm$1.07 & 82.57$\pm$5.43 & 91.76$\pm$1.31 & 98.45$\pm$0.50 & 94.55$\pm$1.80  \\ \midrule
\multirow{4}{*}{HAR Laying} & AUC    & 59.73$\pm$4.76  & 95.30$\pm$2.97 & 95.20$\pm$2.38 & 69.98$\pm$1.94 & \textbf{100}$\pm$0      & 94.60$\pm$5.90 & 62.82$\pm$1.92 & \underline{99.99}$\pm$0       \\
                            & F1     & 42.89$\pm$0.24  & 81.00$\pm$0.13 & 83.73$\pm$2.95 & 50.19$\pm$3.31 & \underline{99.14}$\pm$1.15 & 72.06$\pm$4.54 & 33.69$\pm$3.31 & \textbf{99.35}$\pm$0.20       \\
                            & Prec   & 27.30$\pm$0.19  & 73.97$\pm$0.10 & 77.39$\pm$5.84 & 34.70$\pm$1.62 & 100$\pm$0      & 64.51$\pm$6.73 & 25.47$\pm$1.62 & 98.71$\pm$0.39       \\
                            & Rec    & 100$\pm$0       & 91.56$\pm$0.10 & 91.81$\pm$5.16 & 97.95$\pm$6.96 & 98.32$\pm$2.24 & 84.67$\pm$5.92 & 53.32$\pm$6.96 & 100$\pm$0         \\ \midrule
Aerobic Digit &Avg AUC   & 79.70           &  77.27        & 77.67          & 77.80               & 74.21           & \underline{83.12}          & 81.52          & \textbf{87.70} \\ \midrule
UWave &Avg AUC   & 84.33           & \underline{86.95}              & 85.35          &86.61               & 70.32           & 81.11          & 73.20          & \textbf{88.72}   \\ \bottomrule 
\end{tabular}
}

\end{table*}

\begin{table}[!htbp]
\centering
\caption{Performance of ExDNN variants. We fix the number of exemplars to 32. Best results are highlighted in bold.}
\label{tab:exdnnvariants}
\resizebox{0.5\textwidth}{!}{%
\begin{tabular}{@{}l|cc|cc|cccc|c@{}}
\toprule
Methods        & \multicolumn{2}{c}{2D Gesture}  &AerobicDigit \\ \midrule
               & AUC        & F1                  & Avg AUC      \\
ExDNN w/ AE & 68.83$\pm$3.17 & 61.35$\pm$4.36  & 83.57        \\
ExDNN w/o clus & 85.96$\pm$1.69 & 67.11$\pm$2.04 & 68.57        \\
ExDNN w/o bal   & 85.01$\pm$1.02 & 64.46$\pm$1.05  & 83.67        \\
ExDNN w/o abs   & 83.26$\pm$1.34 & 63.93$\pm$2.43& 85.94        \\
ExDNN          & \textbf{86.31}$\pm$1.47     & \textbf{67.34}$\pm$1.98    &\textbf{87.70}       \\ \bottomrule
\end{tabular}
}
\end{table}

\begin{table}[]
\centering
\caption{Performance of ExDNN with different numbers of exemplars. Best results are highlighted in bold.}
\label{tab:exdnncluster}
\resizebox{0.4\textwidth}{!}{%
\begin{tabular}{@{}l|cc|cc|cccc|c@{}}
\toprule
K & \multicolumn{2}{c}{2D Gesture}&AerobicDigit \\ \midrule
            & AUC        & F1             & Avg AUC      \\
8           & 82.34$\pm$1.60 & 62.30$\pm$3.47  & 75.51        \\
16          & 84.94$\pm$2.88 & 66.15$\pm$3.63 & 84.02        \\
32          & 86.31$\pm$3.55 & 67.34$\pm$4.67 & 87.70        \\
64          & 87.34$\pm$0.72 & 68.41$\pm$1.35 & \textbf{88.62}        \\
128         & \textbf{88.37}$\pm$0.82 & \textbf{71.13}$\pm$2.31 & 88.61        \\ \bottomrule
\end{tabular}
}
\end{table}

\subsubsection{Ablation Studies} \label{sec:ablation}
In this section, we conduct ablation studies on 2D Gesture and AerobicDigit. For AerobicDigit, only averaged AUC is reported. We first study the effectiveness of different components of the proposed ExDNN. The results are shown in Table \ref{tab:exdnnvariants}. We denote ExDNN without the cluster term in Eq. (\ref{eq:localall}) as ExDNN w/o clus,  ExDNN without the balanced term in Eq. (\ref{eq:localall}) as ExDNN w/o bal, and ExDNN without the absolute term in Eq. (\ref{eq:localall}) as ExDNN w/o abs. Moreover, to validate the effectiveness of the proposed DRP for representation, we replace DRP with a pretrained network by auto-encoder as ExDNN w/ AE. The experiments are conducted with 32 exemplars for 2D Gesture and AerobicDigit. Based on the results, we could conclude that the cluster term brought from DEC \cite{xie2016unsupervised} is essential for the success of our method for AerobicDigit whose normal cases are heterogeneous. The proposed absolute term in Eq. (\ref{eq:localall}) could consistently boost the anomaly detection performance on different datasets. The balanced term could also improve the performance and stabilize the training process. Comparing ExDNN with ExDNN w/ AE, DRP is a better choice than the autoencoder for ExDNN. 

\subsubsection{Influence of the number of exemplars}
We study the influence of the number of exemplars, and we detail the results in Table \ref{tab:exdnncluster}. The results indicate that increasing the number of exemplars could improve the performance of ExDNN for heterogeneous normal cases in general. 

\subsubsection{Contamincation Study}
\begin{figure}
\centering
\subfloat[AUC]{
\includegraphics[width=0.4\textwidth]{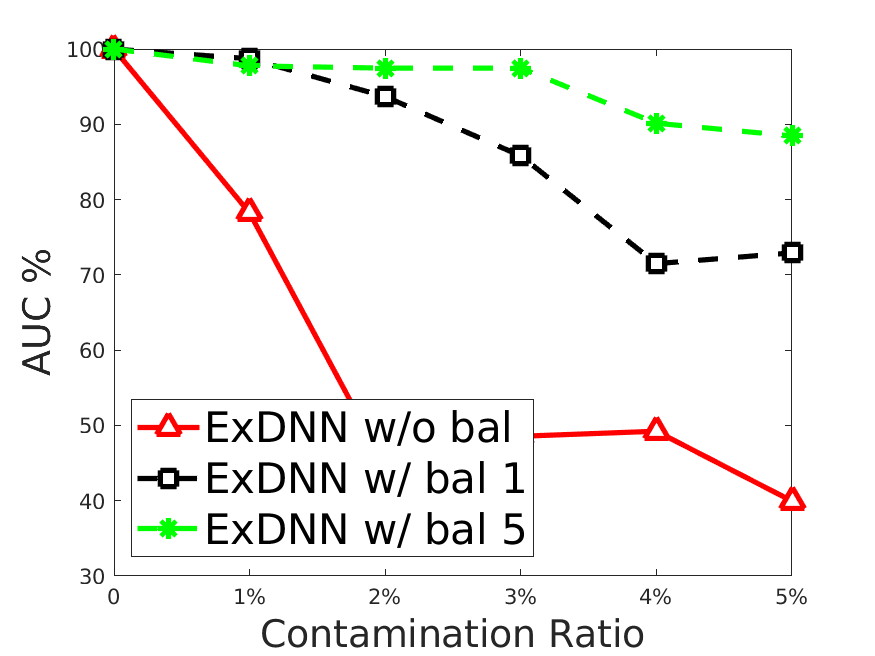}
}
\subfloat[F1]{
\includegraphics[width=0.4\textwidth]{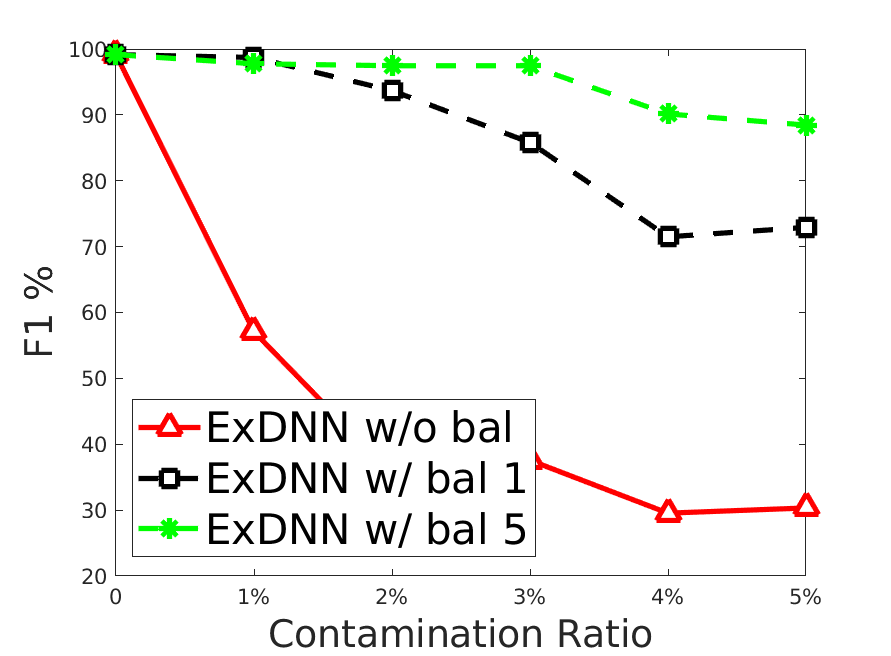}
}
\caption{Anomaly detection performance on the contaminated HAR Laying dataset.}
\label{fig:exdnncontaminated}
\end{figure}

ExDNN is developed based on the assumption that the training set only contains normal data. To study the performance of ExDNN on contaminated data, we inject different percentages of abnormal data into the training set. We vary this percentage from 1\% to 5\% and conduct the experiments on HAR Laying. To validate the effectiveness of the balanced term for handling noisy samples, we vary the weight of the balanced term from $\lbrace 0,1,5\rbrace$, and these variants of ExDNN are denoted as ExDNN w/o bal, ExDNN w/ bal 1, and ExDNN w/ bal 5 respectively. According to the experiments shown in Fig. \ref{fig:exdnncontaminated}, the balanced term can significantly boost the robustness of ExDNN. Although it is better to set a large weight for the balanced term, the default setting of ExDNN (ExDNN w/ bal 1) is already robust to abnormal sample contamination, and is sufficient for most real-life applications.


\subsubsection{Training Time}
We compare the training time of reconstruction-based (LSTM AE), one class-based (Deep SVDD), and clustering-based (ExDNN) deep anomaly detection methods. There is no need to conduct experiments with other deep learning methods as they are developed based on either LSTM AE or Deep SVDD. The results are shown in Table \ref{tab:runnigntime}. LSTM AE is more computational expensive than ExDNN and Deep SVDD for the decoder process, and the auto-encoder pertaining required by Deep SVDD will also significantly slow down its training process. ExDNN replaces the auto-encoder with DRP for representation learning which is effective and brings negligible additional computation cost. 

\begin{table}[]
\centering
\caption{Running time (second) per epoch of three different types of anomaly detection methods. Deep SVDD requires extra epochs for auto-encoder pretraining.}
\label{tab:runnigntime}
\resizebox{0.6\columnwidth}{!}{
\begin{tabular}{lcccc}
\toprule
Methods    & 2D Gesture & SWAT  & ECG5000 & HAR Laying \\ \midrule
ExDNN     & 3.61    & 10.87 & 0.12    & 2.05       \\
Deep SVDD & \textbf{3.41}    & \textbf{10.11} & \textbf{0.10}    & \textbf{1.87}       \\
LSTM AE   & 5.53    & 17.26 & 0.19    & 3.24       \\ \bottomrule
\end{tabular}
}

\end{table}

\begin{table*}[!htbp]
\centering
\caption{Federated anomaly detection performance on six MTS datasets. Best performance is highlighted in bold.}
\label{tab:fedexdnn4}
\resizebox{\textwidth}{!}{
\begin{tabular}{llcccc|cccc}
\toprule
Dataset                     &      & FedAvgAE   & FedProxAE  & FedAvgSVDD  & FedProxSVDD & FedAvgEx   & FedProxEx  & FedKmsEx   & Fed-ExDNN  \\ \hline
\multirow{4}{*}{2D Gesture} & AUC  & 80.64$\pm$0.89 & 81.29$\pm$1.14 & 70.22$\pm$4.61  & 77.64$\pm$4.45  & 82.81$\pm$1.37 & 82.79$\pm$0.27 & 83.57$\pm$2.70 & \textbf{85.24}$\pm$1.20 \\
                            & F1   & 57.92$\pm$1.32 & 58.13$\pm$1.28 & 51.86$\pm$5.74  & 56.58$\pm$2.60  & 63.12$\pm$2.24 & 63.97$\pm$0.93 & 62.56$\pm$1.47 & \textbf{64.75}$\pm$2.41 \\
                            & Prec & 50.55$\pm$2.63 & 50.17$\pm$1.65 & 45.00$\pm$8.12  & 53.85$\pm$2.67  & 57.39$\pm$2.65 & 58.42$\pm$1.79 & 59.29$\pm$0.54 & 59.88$\pm$1.34 \\
                            & Rec  & 65.44$\pm$1.69 & 66.15$\pm$0.84 & 69.44$\pm$9.54  & 59.64$\pm$2.99  & 72.99$\pm$3.25 & 71.08$\pm$4.93 & 66.25$\pm$2.61 & 74.82$\pm$3.11 \\ \midrule
\multirow{4}{*}{SWaT}       & AUC  & 90.19$\pm$0.17 & 90.31$\pm$0.46 & \textbf{91.57}$\pm$2.49  & 90.53$\pm$5.25  & 86.20$\pm$2.96 & 88.47$\pm$1.37 & 84.41$\pm$1.23 & 89.27$\pm$1.91 \\
                            & F1   & 76.94$\pm$0.08 & 76.87$\pm$0.04 & 79.19$\pm$2.32  & 79.76$\pm$4.27  & 80.38$\pm$0.35 & 80.95$\pm$0.81 & 80.58$\pm$0.08 & \textbf{83.61}$\pm$0.58 \\
                            & Prec & 62.52$\pm$1.10 & 62.44$\pm$0.05 & 65.79$\pm$3.33  & 66.75$\pm$6.08  & 70.68$\pm$1.86 & 72.17$\pm$1.88 & 70.63$\pm$0.09 & 72.57$\pm$0.25 \\
                            & Rec  & 100$\pm$0      & 100$\pm$0      & 99.63$\pm$0.27  & 99.54$\pm$0.11  & 93.35$\pm$2.89 & 92.22$\pm$0.97 & 93.81$\pm$0.39 & 94.90$\pm$0.17 \\ \midrule
\multirow{4}{*}{ECG5000}  & AUC  & 94.01$\pm$0.62 & 95.88$\pm$1.87 & 94.81$\pm$4.71  & 95.47$\pm$4.01  & 97.39$\pm$0.79 & 97.36$\pm$0.80 & 96.50$\pm$2.03 & \textbf{98.17}$\pm$0.53 \\
                            & F1   & 89.30$\pm$0.20 & 90.66$\pm$2.24 & 91.83$\pm$4.95  & 92.67$\pm$4.47  & 92.61$\pm$1.49 & 92.61$\pm$1.60 & 91.08$\pm$4.35 & \textbf{93.86}$\pm$0.48 \\
                            & Prec & 82.34$\pm$0.62 & 84.15$\pm$4.27 & 87.47$\pm$9.55  & 88.33$\pm$9.02  & 93.83$\pm$2.39 & 93.94$\pm$2.39 & 92.31$\pm$2.99 & 93.49$\pm$0.78 \\
                            & Rec  & 97.83$\pm$1.02 & 98.43$\pm$0.50 & 97.68$\pm$2.11  & 98.14$\pm$1.21  & 91.45$\pm$1.48 & 91.36$\pm$1.69 & 89.93$\pm$5.64 & 94.00$\pm$0.46 \\ \midrule
\multirow{4}{*}{HAR Laying} & AUC  & 89.74$\pm$1.31 & 88.74$\pm$0.21 & 99.66$\pm$0.09  & 99.55$\pm$0.06  & 99.43$\pm$0.56 & 99.71$\pm$0.22 & 99.99$\pm$0.01 & \textbf{99.99}$\pm$0.01 \\
                            & F1   & 69.52$\pm$2.15 & 67.63$\pm$0.56 & 98.35$\pm$0.54  & 97.38$\pm$1.33  & 97.39$\pm$0.16 & 96.95$\pm$0.47 & 97.87$\pm$0.15 & \textbf{98.99}$\pm$0.08 \\
                            & Prec & 55.13$\pm$2.70 & 52.80$\pm$0.57 & 96.77$\pm$1.05  & 95.09$\pm$2.37  & 97.51$\pm$2.30 & 99.41$\pm$0.59 & 97.30$\pm$1.75 & 98.08$\pm$0.27 \\
                            & Rec  & 94.23$\pm$0.19 & 94.04$\pm$0.37 & 100$\pm$0       & 99.81$\pm$0.19  & 97.39$\pm$2.61 & 94.60$\pm$0.37 & 98.51$\pm$1.49 & 99.91$\pm$0.09 \\ \midrule
AerobicDigit                &Avg AUC  & 63.76      & 65.57      & 66.97      & 68.52       & 67.84      & 70.18      & 72.86      & \textbf{79.53}       \\ \midrule
UWave &Avg AUC   &  81.74         &84.06   & 80.27           &81.43             &81.17            & 81.59           &79.65            &\textbf{86.77}   \\ 
\bottomrule
\end{tabular}
}
\end{table*}

\subsection{Fed-ExDNN for Fedrated Anomaly Detection}
After demonstrating the effectiveness of ExDNN, we conduct experiments to validate the superiority of Fed-ExDNN for FedUAD. 

\subsubsection{Comparison Methods and Experimental Settings}
The experiments are conducted on all six MTS datasets. To simulate federated settings, for 2D Gesture, SWaT, and ECG5000, we sequentially partition the data into $L$ different parts and assign them to $L$ different edge devices. For HAR Laying, we assign the samples from each subject to an edge device and discard samples of two random activities for each subject. For AerobicDigit and UWave, the training set on each edge device is constructed by randomly select 900 and 300 samples from 3 different classes.

We implement several federated anomaly detection baselines. We aggregate the local models trained by LSTM AE (Deep SVDD) by Federated Average (FedAvg) and Federated Proximal (FedProx) as FedAvgAE (FedAvgSVDD) and FedProxAE (FedProxSVDD), respectively. Moreover, to justify the motivation of FedCC, we propose several variants of ExDNN as follows: we apply FedAvg and FedProx on the proposed ExDNN as FedAvgEx and FedProxEx, respectively; we also adopt Kmeans to aggregate the exemplars as a direct counterpart for FedCC termed as FedKmeans. For FedCC and FedKmeans, we adopt FedAvg to aggregate the feature encoder network. 
The hyperparameters of ExDNN for local training are described in the previous section and we search $\gamma_4$ and $\gamma_5$ from $\lbrace 1,5,10 \rbrace$ for FedCC. For FedProx, we search hyperparameters from $\lbrace 10^{-4},10^{-3},$ $10^{-2},10^{-1} \rbrace$. We conduct federated learning for 5 communication rounds which is sufficient for the performance of all federated anomaly detection methods to converge.  We implement the network for FedCC with a three layer multi-layer perceptron with ReLU as activation function. The batchsize for FedCC is set as 256 and the learning rate is 0.005. We initialize the global exemplars with kmeans++ and train FedCC for 500 steps. All methods are implemented in Pysyft~\cite{he2020fedml} and Pytorch~\cite{paszke2019pytorch}
\subsubsection{Results}
\begin{figure*}[t]
\centering
\subfloat[AerobicDigit 3]{
\includegraphics[width=0.45\textwidth]{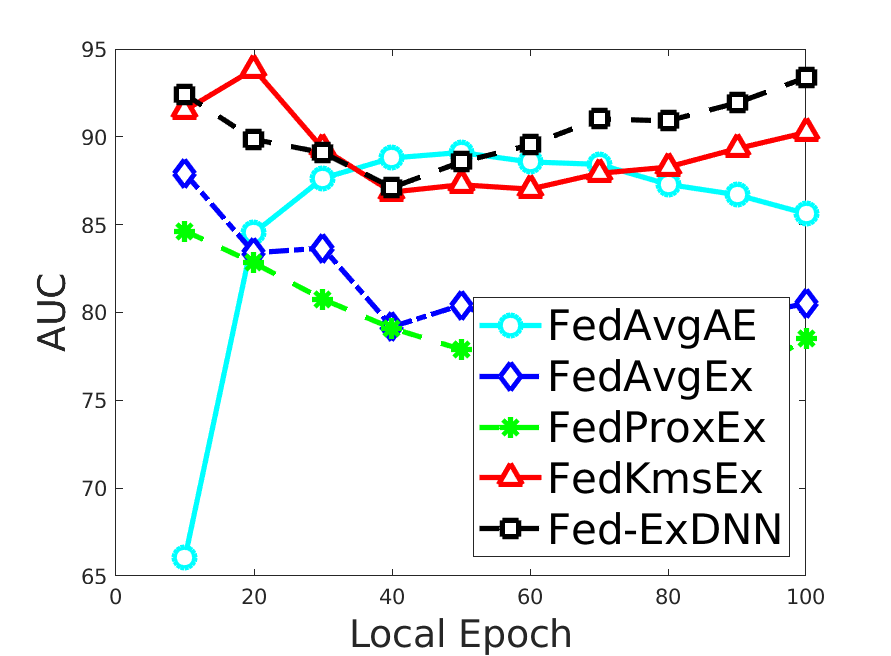}
}
\subfloat[AerobicDigit 6]{
\includegraphics[width=0.45\textwidth]{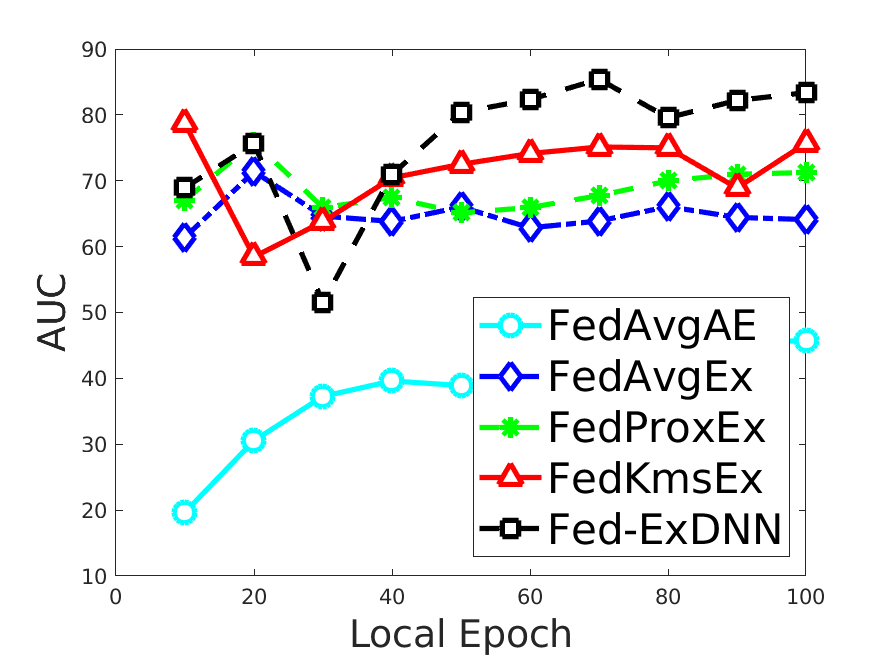}
}\\
\subfloat[Uwave 4]{
\includegraphics[width=0.45\textwidth]{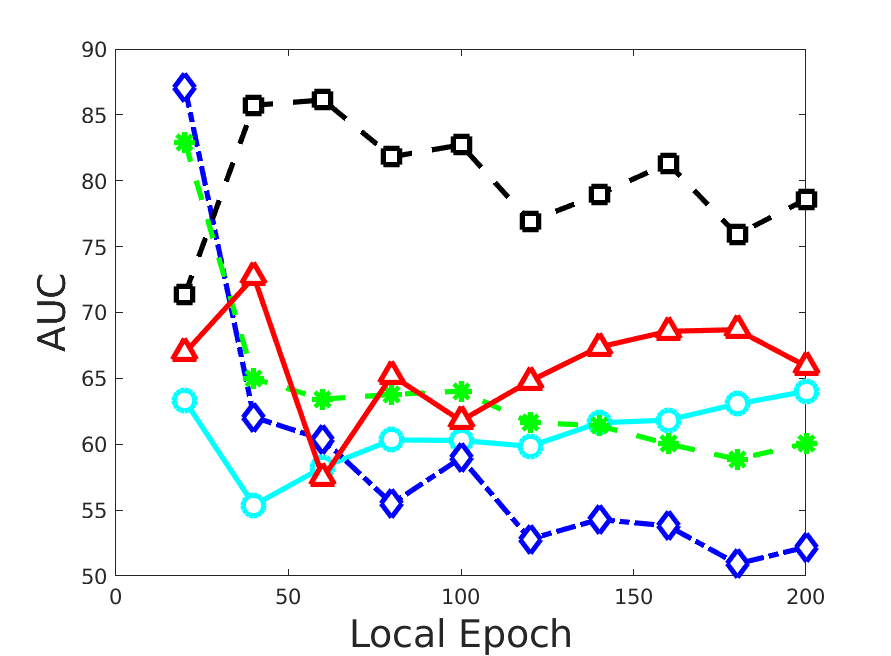}
}
\subfloat[Uwave 5]{
\includegraphics[width=0.45\textwidth]{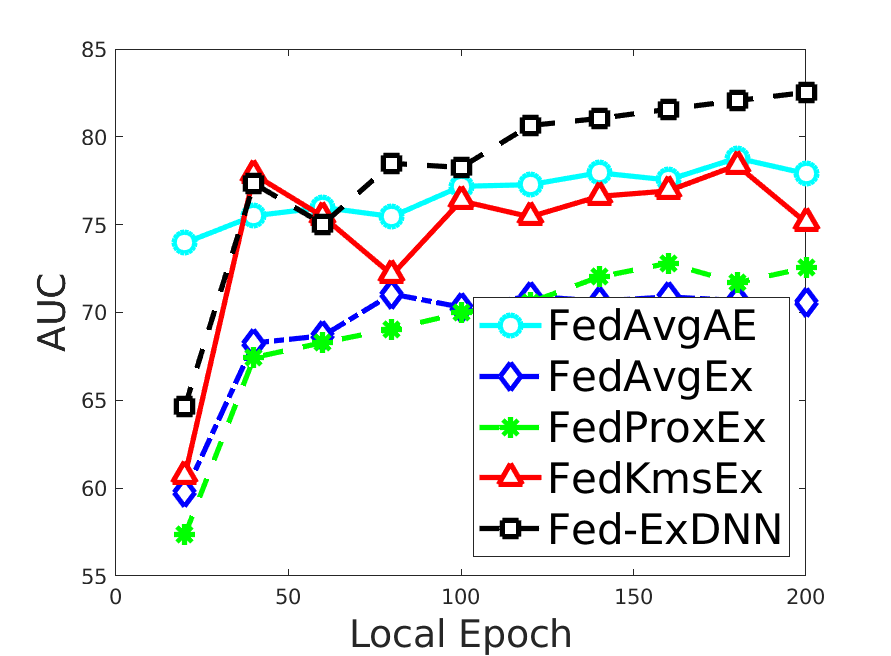}
}

\caption{We show the federated learning results on AerobicDigit and Uwave datasets with different communication round. The X-axis denotes the local training epoch and each point denotes a communication round. We conduct experiments on subsets of AerobicDigit and Uwave for simplicity.}
\label{fig:fedresult}
\end{figure*}

The results of FedUAD are shown in Table \ref{tab:fedexdnn4}. According to the results, overall, the variants of Fed-ExDNN outperform federated anomaly detection baselines. This should be attributed to the fact that ExDNN explicitly takes the heterogeneous data on edge devices into consideration. Moreover, FedKmsEx and Fed-ExDNN outperform FedAvgEx and FedProxEx, since FedKmsEx and Fed-ExDNN could handle the deviation of exemplars. Finally, the proposed Fed-ExDNN performs better than other variants of Federated ExDNN since FedCC can simultaneously learn to align and aggregate local exemplars. Fig. \ref{fig:fedresult} shows the federated learning results for each communication round, and we could conclude that the proposed Fed-ExDNN consistently outperforms its counterparts.

\begin{figure*}[t]
\centering
\subfloat[FedAvgEx]{
\includegraphics[width=0.45\textwidth]{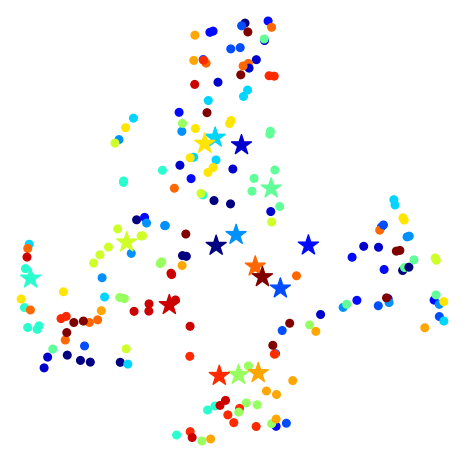}
}
\subfloat[FedKmsEx]{
\includegraphics[width=0.45\textwidth]{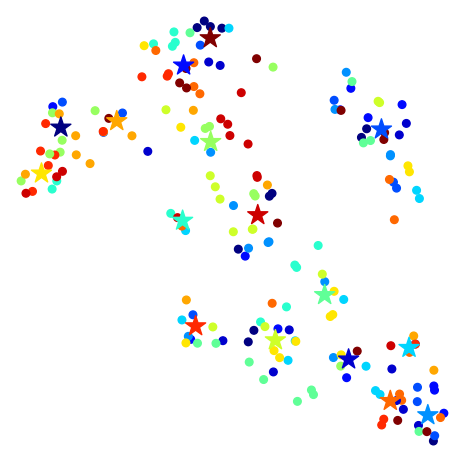}
}\\
\subfloat[FedProxEx]{
\includegraphics[width=0.45\textwidth]{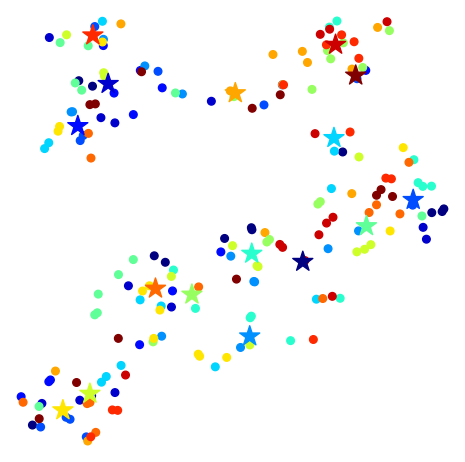}
}
\subfloat[Fed-ExDNN]{
\includegraphics[width=0.45\textwidth]{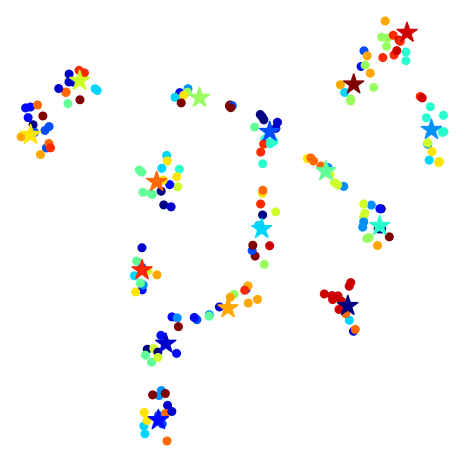}
}
\caption{t-SNE visualization of global exemplar modules aggregated by different methods on AerobicDigit 1. The small circles are local exemplars and the stars are generated global exemplars. Local exemplars with same color means they are at same position slot of different devices, and will be aggregated by FedAvg for a global exemplar with same color.}
\label{fig:fedexdnnvis}
\end{figure*}

\subsubsection{Interpretation}
We provide an intuitive visualization of different global exemplars (stars) learned by different aggregation methods in Fig. \ref{fig:fedexdnnvis}. We could see that the global exemplars generated by FedAvgEx are collapsed. FedProxEx and FedKmsEx neglect the alignment between local exemplars. On the contrary, Fed-ExDNN could produce diverse global exemplars while preserving the alignment to a certain degree. The visualization shows Fed-ExDNN works better than its counterparts.

\section{Conclusions}
In this paper, we developed the Federated Exemplar-based Deep Neural Network (Fed-ExDNN) to perform federated anomaly detection with multivariate time series data. We first investigated the problem of federated unsupervised anomaly detection with multivariate time series data. Then, we developed an Exemplar-based Deep Neural Network (ExDNN) to learn local time series representations based on their compatibility with an exemplar module that can capture varieties of normal patterns. Meanwhile, we also introduced a constrained clustering mechanism to align and aggregate the parameters of local exemplar modules to obtain a unified global exemplar module.
Finally, the updated embedding network (for feature encoding) along with the global exemplar module are sent back to edge devices and the anomaly detection is conducted by comparing to those learned global exemplars. Our thoroughly empirical studies on six public datasets have validated the effectiveness of the proposed ExDNN and Fed-ExDNN.

\begin{appendices}
\section{Additional Results}
We provide more detailed results for AerobicDigit and UWave. For these two datasets, we only report AUC score. Table \ref{tab:abl_components} show the influence of different terms on AerobicDigit. Table \ref{tab:ablexemplars} show how the number of exemplars influences the performance. Table \ref{tab:exdnn2} and Table \ref{tab:fedexdnn2} show the local and federated learning performance. 
\begin{table*}[h]
\centering
\caption{AUC of variants of ExDNN on AerobicDigit.}
\label{tab:abl_components}
\resizebox{0.8\textwidth}{!}{
\begin{tabular}{@{}lcccccccccccc@{}}
\toprule
Method                     & 0          & 1          & 2          & 3          & 4          & 5          & 6           & 7          & 8          & 9     & Avg     \\ \midrule
ExDNN32 w/o indvi & 67.66 & 92.66 & 47.57 & 97.14 & 57.34 & 94.95 & 57.03  & 25.63  & 83.78 & 61.97  &68.57\\
ExDNN32 w/o abs  & 74.48 & 98.77 & 73.37 & 90.73  & 88.29  & 94.74& 73.58   & 88.61 & 96.88 & 79.96 &85.94\\
ExDNN32           & 73.18 & 98.80 & 79.07 & 96.38& 91.42& 95.66 & 75.51& 87.22& 97.98& 81.75 &87.70\\ \bottomrule
\end{tabular}
}

\vspace{3mm}
\centering
\caption{AUC with different number of exemplars on AerobicDigit.}
\label{tab:ablexemplars}
\resizebox{0.8\textwidth}{!}{
\begin{tabular}{@{}lccccccccccc@{}}
\toprule
\# exemplars             & 0     & 1     & 2     & 3     & 4     & 5     & 6     & 7     & 8     & 9     & Avg   \\ \midrule
8   & 71.44 & 98.14 & 77.99 & 97.37 & 79.49 & 94.26 & 65.09 & 90.16 & 96.50 & 62.95 & 83.34 \\
16  & 70.39 & 98.61 & 64.20 & 96.51 & 86.49 & 96.20 & 67.31 & 86.64 & 97.34 & 67.51 & 84.02 \\
32  & 73.61 & 99.04 & 78.99 & 97.19 & 91.65 & 95.77 & 77.56 & 88.87 & 98.40 & 78.41 & 87.95 \\
64  & 74.19 & 99.30 & 80.56 & 97.05 & 90.78 & 96.28 & 79.58 & 91.59 & 98.73 & 78.17 & 88.62 \\
128 & 72.06 & 99.34 & 78.11 & 97.27 & 92.68 & 97.19 & 79.87 & 91.39 & 98.86 & 79.35 & 88.61 \\ \bottomrule
\end{tabular}
}
\end{table*}

\begin{table*}[h]
\centering
\caption{Unsupervised Anomaly Detection performance (AUC) on AerobicDigit and UWave. Best AUC is highlighted in bold.}
\resizebox{\textwidth}{!}{
\begin{tabular}{@{}lcccccccc@{}}
\toprule
Class     & LSTM AE         & BeatGAN        & MemAE          & USAD           & Deep SVDD       & CVDD           & DAGMM          & Ours           \\ \midrule
\multicolumn{9}{c}{AerobicDigit}                                                                                                                               \\
0     & 70.19$\pm$2.28  & 65.79$\pm$5.23 & 64.84$\pm$2.72 & 65.25$\pm$1.86 & 70.25$\pm$1.33  & 71.39$\pm$2.20 & \textbf{74.27}$\pm$5.93 & 73.18$\pm$0.38 \\
1     & 97.03$\pm$0.79  & 94.86$\pm$0.79 & 98.35$\pm$0.05 & 97.12$\pm$0.53 & 94.45$\pm$1.08  & 97.17$\pm$2.50 & 87.00$\pm$1.62 & \textbf{98.80}$\pm$0.26 \\
2      & 74.59$\pm$4.51  & 73.12$\pm$6.00 & 63.02$\pm$6.65 & 69.62$\pm$5.18 & 69.28$\pm$7.68  & 76.94$\pm$6.35 & \textbf{79.70}$\pm$1.51 & 79.07$\pm$0.78 \\
3      & 92.51$\pm$3.14  & 92.29$\pm$1.29 & 93.97$\pm$0.45 & 93.64$\pm$2.32 & 71.05$\pm$1.80  & 96.53$\pm$0.61 & 72.97$\pm$7.40 & \textbf{96.38}$\pm$0.80 \\
4    & 89.18$\pm$1.01  & 85.19$\pm$0.58 & \textbf{92.54}$\pm$2.71 & 84.72$\pm$2.68 & 68.34$\pm$8.43  & 85.52$\pm$3.21 & 74.91$\pm$4.79 & 91.42$\pm$1.49 \\
5     & 91.75$\pm$2.03  & 85.17$\pm$5.36 & 87.17$\pm$3.14 & 84.18$\pm$0.65 & 82.49$\pm$7.53  & 85.25$\pm$4.32 & 83.22$\pm$4.04 & \textbf{95.66}$\pm$0.56 \\
6     & 56.78$\pm$3.21  & 54.65$\pm$3.95 & 55.87$\pm$7.60 & 50.06$\pm$1.06 & 72.09$\pm$11.30 & 75.61$\pm$4.55 & \textbf{84.59}$\pm$5.47 & 75.51$\pm$2.78 \\
7     & 72.60$\pm$0.96  & 64.56$\pm$2.93 & 72.86$\pm$1.63 & 72.33$\pm$5.02 & 72.41$\pm$6.68  & 82.46$\pm$3.23 & 86.08$\pm$0.80 & \textbf{87.22}$\pm$1.28 \\
8     &\textbf{99.20}$\pm$0.08  & 98.81$\pm$0.45 & 98.76$\pm$0.17 & 98.73$\pm$0.36 & 82.83$\pm$1.30  & 87.48$\pm$5.87 & 84.14$\pm$2.73 & 97.98$\pm$0.21 \\
9     & 53.21$\pm$1.34  & 58.21$\pm$2.36 & 49.40$\pm$4.13 & 62.36$\pm$1.25 & 58.99$\pm$11.51 & 72.86$\pm$5.68 & \textbf{88.30}$\pm$4.04 & 81.75$\pm$0.08 \\ \midrule
\multicolumn{9}{c}{UWave}                                                                                                                                      \\
0    & 87.63$\pm$0.29  & 85.73$\pm$8.72 & 78.18$\pm$2.38 & 80.83$\pm$5.56 & 64.48$\pm$7.14  & 72.54$\pm$5.62 & 60.35$\pm$9.58 & \textbf{88.88}$\pm$2.60 \\
1     & \textbf{95.81}$\pm$2.36  & 95.79$\pm$1.59 & 94.36$\pm$1.17 & 95.58$\pm$1.32 & 60.53$\pm$0.96  & 82.76$\pm$4.57 & 50.87$\pm$9.27 & 88.48$\pm$4.06 \\
2     & 94.91$\pm$1.04  & 94.34$\pm$0.65 & \textbf{95.98}$\pm$0.83 & 95.23$\pm$0.55 & 77.57$\pm$4.76  & 76.54$\pm$4.63 & 88.21$\pm$2.05 & 92.92$\pm$1.83 \\
3      & 82.40$\pm$1.04  & 85.78$\pm$0.47 & 85.88$\pm$1.05 & 86.50$\pm$2.56 & 78.75$\pm$0.83  & 81.25$\pm$4.32 & 86.52$\pm$2.70 & \textbf{89.21}$\pm$1.92 \\
4     & 53.93$\pm$10.06 & 73.03$\pm$8.66 & 79.03$\pm$0.96 & 69.15$\pm$6.11 & \textbf{86.21}$\pm$2.96  & 83.26$\pm$8.54 & 80.23$\pm$4.40 & 72.66$\pm$1.47 \\
5     & 80.42$\pm$0.77  & 77.50$\pm$8.51 & 72.86$\pm$1.04 & 80.43$\pm$3.78 & 80.51$\pm$3.98  & 81.24$\pm$3.51 & 78.40$\pm$8.64 & \textbf{84.12}$\pm$1.81 \\
6     & 86.74$\pm$5.56  & 87.29$\pm$1.16 & 79.63$\pm$6.50 & 88.19$\pm$6.81 & 66.89$\pm$17.77 & 89.16$\pm$5.31 & 64.25$\pm$5.42 & \textbf{95.85}$\pm$2.07 \\
7   & 92.81$\pm$4.40  & 96.14$\pm$0.19 & 96.88$\pm$0.53 & 96.96$\pm$0.73 & 47.64$\pm$9.78  & 82.15$\pm$6.31 & 76.83$\pm$7.91 & \textbf{97.64}$\pm$0.41 \\ 
 \bottomrule
\end{tabular}
}
\label{tab:exdnn2}
\end{table*}

\begin{table*}[h]
\centering

\caption{Federated Anomaly Detection on AerobicDigit and UWave. Best Results are highlighted in bold.}
\resizebox{\textwidth}{!}{
\begin{tabular}{lcccc|cccc}
\hline
Class & FedAvgAE   & FedProxAE  & FedAvgSVDD & FedProxSVDD & FedAvgEx   & FedProxEx  & FedKmeans  & Fed-ExDNN   \\ \hline
\multicolumn{9}{c}{AerobicDigit}                                                                                \\
0     & 54.79$\pm$1.02 & 55.03$\pm$1.56 & 53.63$\pm$2.13 & 56.98$\pm$4.68  & 63.45$\pm$1.95 & 63.89$\pm$1.14 & 56.87$\pm$1.58 & \textbf{67.67}$\pm$8.28  \\
1     & 92.52$\pm$1.03 & 92.12$\pm$1.37 & 77.74$\pm$1.48 & 80.96$\pm$2.28  & 93.17$\pm$0.51 & 90.33$\pm$2.83 & 91.11$\pm$1.72 & \textbf{96.34}$\pm$0.55  \\
2     & 37.18$\pm$4.54 & 40.78$\pm$4.34 & 60.94$\pm$5.14 & 63.20$\pm$11.80 & 60.49$\pm$6.20 & 65.28$\pm$2.92 & 59.83$\pm$0.80 &\textbf{ 79.87}$\pm$4.97  \\
3     & 76.75$\pm$0.39 & 81.34$\pm$0.22 & 72.38$\pm$2.72 & 73.16$\pm$5.42  & 75.97$\pm$5.94 & 78.41$\pm$5.71 & 84.47$\pm$0.32 & \textbf{82.19}$\pm$0.31  \\
4     & 81.42$\pm$3.87 & 79.37$\pm$2.05 & 69.49$\pm$3.80 & 71.20$\pm$1.86  & 73.40$\pm$3.69 & 73.16$\pm$2.20 & 65.53$\pm$9.33 & \textbf{82.41}$\pm$5.30  \\
5     & 71.22$\pm$3.00 & 78.04$\pm$0.49 & 55.97$\pm$2.81 & 58.91$\pm$1.90  & 71.29$\pm$6.60 & 71.88$\pm$3.11 & 79.40$\pm$3.16 & \textbf{80.18}$\pm$2.66  \\
6     & 40.58$\pm$0.10 & 42.05$\pm$4.08 & 72.87$\pm$4.63 & 75.07$\pm$2.25  & 61.54$\pm$3.19 & 63.66$\pm$2.12 & 73.10$\pm$3.57 & \textbf{81.31}$\pm$10.03 \\
7     & 59.86$\pm$1.26 & 60.81$\pm$2.29 & 70.78$\pm$6.60 & 72.38$\pm$6.08  & 68.92$\pm$0.44 & 75.82$\pm$2.42 & 73.62$\pm$3.03 & \textbf{78.62}$\pm$1.70  \\
8     & 91.22$\pm$0.95 & \textbf{92.67}$\pm$0.59 & 72.14$\pm$5.38 & 69.06$\pm$6.06  & 69.27$\pm$8.07 & 77.95$\pm$6.35 & 88.96$\pm$2.89 & 88.44$\pm$0.54  \\
9     & 32.02$\pm$6.54 & 33.44$\pm$6.12 & 63.82$\pm$3.08 & \textbf{64.29}$\pm$3.46  & 40.94$\pm$3.54 & 41.48$\pm$8.53 & 55.71$\pm$8.65 & 58.31$\pm$4.81  \\ \hline
\multicolumn{9}{c}{UWave}                                                                                       \\
0     & 82.08$\pm$2.40 & 80.83$\pm$0.63 & 63.91$\pm$6.07 & 70.72$\pm$4.26  & 79.51$\pm$5.38 & 74.36$\pm$1.79 & 72.05$\pm$8.66 & \textbf{86.76}$\pm$4.34  \\
1     & 89.75$\pm$4.58 & 87.26$\pm$2.82 & 93.44$\pm$5.31 & 89.46$\pm$3.61  & 84.59$\pm$1.14 & 85.73$\pm$2.66 & 87.73$\pm$2.79 & \textbf{92.09}$\pm$1.52  \\
2     & 90.32$\pm$2.68 & \textbf{91.18}$\pm$2.63 & 88.40$\pm$1.44 & 84.71$\pm$6.47  & 84.49$\pm$2.34 & 81.71$\pm$0.29 & 84.18$\pm$4.12 & 87.59$\pm$1.57  \\
3     & 78.56$\pm$4.55 & 84.97$\pm$4.38 & 84.50$\pm$1.78 & \textbf{86.80}$\pm$2.40  & 77.17$\pm$1.58 & 81.73$\pm$2.27 & 80.34$\pm$2.51 & 85.85$\pm$4.97  \\
4     & 62.43$\pm$0.84 & 76.59$\pm$2.03 & 74.21$\pm$2.40 & 76.91$\pm$1.24  & 76.92$\pm$3.46 & 78.08$\pm$0.66 & 65.96$\pm$3.90 & \textbf{81.54}$\pm$2.48  \\
5     & {79.74}$\pm$3.13 & 78.95$\pm$2.07 & 69.28$\pm$2.51 & 73.30$\pm$0.60  & 67.66$\pm$4.87 & 70.51$\pm$1.28 & 73.05$\pm$3.73 &\textbf{ 82.20}$\pm$2.92  \\
6     & 79.71$\pm$0.20 & 78.40$\pm$3.54 & 75.77$\pm$4.40 & 78.27$\pm$0.98  & 85.90$\pm$2.47 & 86.36$\pm$0.30 & 82.76$\pm$3.38 & \textbf{89.57}$\pm$4.05  \\
7     & 91.39$\pm$3.76 & \textbf{94.30}$\pm$2.54 & 92.66$\pm$3.13 & 91.34$\pm$0.45  & 93.14$\pm$1.28 & 94.25$\pm$0.79 & 91.15$\pm$2.22 & 91.58$\pm$5.19  \\ \bottomrule

\end{tabular}
}
\label{tab:fedexdnn2}
\end{table*}

\end{appendices}
\bibliography{fed}
\end{document}